\documentclass[sigconf,screen]{acmart}
\AtBeginDocument{%
  }

\setcopyright{rightsretained}

\copyrightyear{2025}
\acmYear{2025}
\acmDOI{10.1145/3721146.3721956}

\acmConference[EuroMLSys '25]{5th Workshop
on Machine Learning and Systems}{March 30-April 3, 2025}{Rotterdam, Netherlands}
\acmISBN{979-8-4007-1538-9/25/03}

\usepackage{amsmath}
\usepackage{subcaption}
\usepackage{algorithmicx}
\usepackage[noend]{algpseudocode}
\usepackage{algorithm}
\newcommand{\ie}{\textit{i.e.}}
\newcommand{\eg}{\textit{e.g.}}
\usepackage{enumitem}
\setlist{nolistsep}

\usepackage{tikz}

\begin{document}

\title[Fine-Grained Preemption and Priority-Aware Scheduling in LLMs]{Priority-Aware Preemptive Scheduling for Mixed-Priority Workloads in MoE Inference}

\author{Mohammad Siavashi}
\affiliation{%
  \institution{KTH Royal Institute of Technology}
  \city{Stockholm}
  \country{Sweden}
}
\orcid{0000-0002-2600-9025}

\author{Faezeh Keshmiri Dindarloo}
\authornote{Work done while at KTH Royal Institute of Technology}
\affiliation{%
  \institution{Unaffiliated Researcher}
  \city{Stockholm}
  \country{Sweden}
}
\orcid{0009-0008-2161-0437}

\author{Dejan Kostić}
\affiliation{%
  \institution{KTH Royal Institute of Technology}
  \city{Stockholm}
  \country{Sweden}
}
\orcid{0000-0002-1256-1070}

\author{Marco Chiesa}
\affiliation{%
  \institution{KTH Royal Institute of Technology}
  \city{Stockholm}
  \country{Sweden}
}
\orcid{0000-0002-9675-9729}

\renewcommand{\shortauthors}{Mohammad et al.}

\begin{abstract}
    Large Language Models have revolutionized natural language processing, yet serving them efficiently in data centers remains challenging due to mixed workloads comprising latency-sensitive (LS) and best-effort (BE) jobs. Existing inference systems employ iteration-level first-come-first-served scheduling, causing head-of-line blocking when BE jobs delay LS jobs. We introduce QLLM, a novel inference system designed for Mixture of Experts (MoE) models, featuring a fine-grained, priority-aware preemptive scheduler. QLLM enables expert-level preemption, deferring BE job execution while minimizing LS time-to-first-token (TTFT). Our approach removes iteration-level scheduling constraints, enabling the scheduler to preempt jobs at any layer based on priority. Evaluations on an Nvidia A100 GPU show that QLLM significantly improves performance. It reduces LS TTFT by an average of $65.5\times$ and meets the SLO at up to $7$ requests/sec, whereas the baseline fails to do so under the tested workload. Additionally, it cuts LS turnaround time by up to $12.8\times$ without impacting throughput. QLLM is modular, extensible, and seamlessly integrates with Hugging Face MoE models.

\end{abstract}

\ccsdesc[500]{Software and its engineering~Software performance}
\ccsdesc[500]{Computing methodologies~Neural networks}




\keywords{Large Language Models, Mixture-of-Experts, Preemptive Scheduling, Latency-Sensitive Inference, GPU Acceleration, Priority-Aware Scheduling}



\maketitle

\section{Introduction}
Large Language Models (LLMs) have significantly advanced the domain of Natural Language Processing (NLP), enabling tasks such as machine translation, summarization \cite{nallapati2016abstractive, paulus2018deep, see2017get}, code synthesis and completion \cite{chen2021evaluating}, and conversational AI \cite{adiwardana2020towards, roller2021recipes}. The Mixture of Experts (MoE) architecture \cite{jiang2024mixtral}, a transformer variant that selectively activates subsets of specialized feedforward layers per token, has emerged as the preferred paradigm for large-scale models that can attain superior performance while ensuring rapid inference. Although the model in its entirety can be extensive (e.g., 671 B parameters for DeepSeek-R1 \cite{deepseek2025r1}), the selective activation of so-called experts can substantially reduce inferencing costs and subsequently enhance adoption \cite{krajewski2024scaling}.


Data centers serving LLMs typically handle two types of requests: high priority, which are \textit{latency-sensitive (LS)}, and low priority, which are \textit{best-effort (BE)} \cite{sun2024llumnix,shen2024fastswitch}. High priority requests might originate from users with paid subscriptions that include a \textit{Service Level Objective} (SLO) agreement or from interactive applications like ChatBots. In contrast, BE requests could come from users on free tiers or throughput-oriented jobs such as document summarization \cite{sun2024llumnix}. Consequently, inference systems must identify and differentiate between LS and BE requests, ensuring low \textit{time-to-first-token (TTFT)} and quick turnaround time for LS jobs while maintaining high throughput for BE jobs.

Current large-scale inference systems for LLMs, such as Orca~\cite{jeon2023orca}, vLLM~\cite{shi2023vllm}, and Hugging Face (HF) TGI~\cite{tgi}, predominantly employ \textit{iteration-level scheduling}, where new jobs are incorporated, and completed jobs are removed only at the end of each iteration. While this approach enables efficient batching, it adheres to a \textit{first-come-first-served (FCFS)} strategy, treating all inference jobs equally and failing to prioritize LS jobs over BE workloads. As a result, LS jobs frequently experience \textit{head-of-line (HOL) blocking}~\cite{kaffes2019shinjuku}, where large BE jobs with long input and output sequences monopolize resources, delaying LS execution. These delays are particularly pronounced in LLM inference workloads, where request sizes vary significantly, exacerbating scheduling inefficiencies. Addressing this issue requires an inference system capable of distinguishing between LS and BE jobs, reacting to LS job arrivals with minimal delay, and enabling \textit{fine-grained preemption} of BE computations to improve overall system responsiveness.

The dynamic top-\textit{k} token routing inherent to MoE architectures necessitates granular state management: preempted sequences must retain not only their KV cache, but also expert assignments, routing metadata, and partial computations of the \textit{k} selected experts to ensure deterministic resumption \cite{shazeer2017outrageously}. Fine-grained preemptive scheduling is becoming increasingly feasible with modern hardware advancements, such as NVLink’s low-latency interconnects \cite{nvidia_nvlink} and unified memory architectures \cite{nvidia_grace_unified_memory}, which are becoming increasingly prevalent in data centers. However, these advancements also require specialized scheduling mechanisms tailored to contemporary MoE models.

This paper introduces QLLM, an inference system that reduces LS job latency in MoE models through fine-grained preemption and priority-aware scheduling at the expert level. QLLM features: (1) a redesigned MoE layer with per-expert queues for dynamic buffering and low-overhead state management, and (2) a priority-aware scheduler that mitigates HOL blocking by distinguishing LS and BE jobs. Unlike existing inference systems using iteration-level execution, QLLM allows independent expert processing, allowing LS jobs to preempt BE jobs without discarding intermediate computations. An efficient state management mechanism preserves execution progress, allowing seamless BE resumption. The scheduler optimizes LS latency while maintaining high throughput.


Our evaluation on an Nvidia A100 80 GB GPU shows that QLLM reduces TTFT by up to $101.6\times$ (avg. $65.2\times$), enabling SLO compliance for up to 7 jobs per second.  QLLM maintains comparable or higher throughput than existing systems and reduces LS turnaround time by up to $12.8\times$.

Our work makes the following contributions:

\noindent \textbf{Novel MoE Layer Design:} We introduce per-expert queues to enable token buffering and deferred execution, eliminating rigid layer-wise synchronization. This design allows independent expert execution, enhancing scheduling flexibility.

\noindent \textbf{Priority-Aware Scheduler:} QLLM incorporates a scheduler that differentiates LS and BE jobs, ensuring low-latency scheduling and efficient GPU resource allocation.

\noindent \textbf{Fine-Grained Expert-Level Preemption:} QLLM enables BE job preemption at the expert level, reducing LS job queuing delays. This is achieved via a lightweight state management mechanism, a unified KV cache abstraction for batch updates, and per-expert queuing.

\noindent \textbf{Real-World Evaluation:} We evaluate QLLM on real hardware with Mixtral 8×7B, demonstrating improved LS job latency while maintaining high throughput.

\noindent \textbf{Modular and Extensible Framework:} QLLM integrates seamlessly with Hugging Face MoE models with minimal modifications (e.g., class inheritance), facilitating deployment, extensibility, and further research in MoE inference.

This paper discusses our initial findings, a preliminary evaluation, and limitations. Our ultimate plan is to release QLLM as an open-source project in future versions.

\section{Background and Motivation}

\subsection{Mixture of Experts Models}

MoE models are a type of transformer model \cite{vaswani2017attention} designed to enhance computational efficiency in LLMs by selectively activating only a subset of parameters during inference. Unlike dense transformer models, which process all tokens through fully activated feedforward layers, MoE replaces these layers with multiple expert networks and a router that assigns each token to the most relevant experts. This selective activation reduces computational overhead while preserving the overall capacity of the model \cite{shazeer2017outrageously,krajewski2024scaling}. State-of-the-art MoE models, such as Mixtral \cite{jiang2024mixtral}, OpenAI GPT-4 \cite{berges2024memory}, and DeepSeek v3 \cite{dai2024deepseek}, exemplify this architecture.

\subsection{Prefill and Decode Phases}

In an LLM, each transformer layer's self-attention mechanism determines token interactions using key (K) and value (V) tensors. To generate new tokens efficiently, the model stores KV pairs of all previous tokens in a KV cache, reducing redundant computation and improving inference speed \cite{cachegen,lee2024infinigen}. The output of self-attention is then passed to a per-layer router, which selects experts responsible for generating the final output.

LLM inference consists of two phases: \textit{prefill} and \textit{decode}. Prefill processes input tokens in parallel, generating KV cache entries while producing a single output token. Decode operates iteratively, generating one token at a time while leveraging and updating the KV cache. Prefill is compute-intensive due to self-attention across all tokens, whereas decode is memory-intensive as it computes self-attention only between the new token and previous ones \cite{agrawal2024sarathi,zhong2024distserve,jeon2023orca}.



\subsection{Challenges in Existing Inference Systems}

Modern LLM inference systems employ iteration-level scheduling \cite{nvidia2023fastertransformer,jeon2023orca,shi2023vllm}, where a batch of jobs (\ie, prompts) is processed together, generating one token per job per iteration. An iteration corresponds to a full execution of all model layers. Building upon this, continuous batching, as implemented in vLLM and HF TGI, dynamically updates batch composition at every iteration. It removes completed jobs and adds new ones to maintain batch efficiency. If new jobs arrive and a decode batch has available space, the scheduler stops decoding, prefills the new arrivals, and extends the batch for the next decode iteration \cite{IglesiasGoyanes2024}. Then, the scheduler runs the decode batch in a run-to-completion fashion, meaning iterations continue until the full response of a job is generated.



Existing inference engines typically pad and concatenate input tensors from multiple jobs into a single tensor before execution. Although this improves computation efficiency in run-to-completion systems, it complicates the isolation and updating of individual jobs within inner layers. Since model layers only see low-level tensors, oblivious to corresponding sequences, tracking state updates for each sequence becomes inherently difficult and expensive; thus, state updates for sequences occur only at the iteration level.

For instance, if such a layer-level scheduler operates with a defined policy that preempts tasks based on memory limitations in inner layers, it necessitates the extraction of the token from the running batch tensors. This extraction demands costly structural transformation operations on the tensors to accurately retrieve and preserve the token's state such as kv cache entries, attention mask, hidden states, residuals, routing information, and associated metadata. Additionally, when restoring a preempted token, the system must not only reload its exact state, but also reconfigure its tensors (\eg, by padding the KV cache) to ensure seamless integration with the current running batch, a process susceptible to shape mismatches. Moreover, since current inference systems do not track individual sequences within inner layers (\ie, models only see tensors internally), dynamically modifying batch composition can disrupt data flow, resulting in outputs that no longer align with their corresponding inputs. Such challenges render traditional inference systems impractical for achieving fine-grained scheduling at the layer or expert level. 


\subsection{Limitations of Preemption in Current Systems}

Current inference systems are priority-oblivious. This, combined with the run-to-completion approach of the schedulers, causes delays for LS jobs as they await the completion of time-consuming BE jobs. As a result, the queueing time for LS jobs increases, considerably prolonging TTFT and \textit{turnaround time}, the duration from when a job enters the system until its response is fully generated. This issue is known as HOL blocking in the scheduling context, where extended processing of BE jobs influences the latency of LS jobs \cite{wu2023fastserve, kaffes2019shinjuku}. 

In iteration-level scheduling, there exists a potential for systems to decide on preemption at the granularity of token generation \cite{wu2023fast}. Nevertheless, the increasing number of layers and the strong inter-layer dependencies inherently introduce significant delays until the subsequent iteration. In our experiments, conducted on the A100 with the Mixtral 8x7B model, each decode iteration requires 300-400\,ms. For instance, if a BE job is processing at the first layer when an LS job arrives, the entire iteration execution time will be appended to the LS job's TTFT until it has the opportunity to be scheduled in the next iteration.


An efficient scheduling strategy should have low overhead and support fine-grained job preemption and context switching. It also needs to dynamically prioritize LS jobs to reduce TTFT without significantly affecting system throughput. A strong solution should meet these challenges with minimal model changes, ensuring compatibility across frameworks.

\section{QLLM Design Overview}

Effectively handling mixed-priority workloads requires rethinking the scheduling and inference engine components with the goal of enabling rapid preemption of jobs and context switching at the granularity of experts.

\subsection{Challenges and Design Decisions}
We set the following objectives when developing QLLM.

\vspace{.05in}
\noindent \textbf{Generic and low overhead preemption mechanism.} Fine-grained control of the system state (\eg, KV caches, hidden states) within the inner layers and orchestration of the execution flow is a generic need in LLM serving systems. We use a centralized state management mechanism for sequences and batches that enables real-time tracking and updating of sequence states within layers, avoiding delayed updates at the iteration level.

\noindent \textbf{Minimizing LS Job latency without sacrificing throughput.} The key challenge is reacting to high-priority workloads with fine-grained timing. To address this, QLLM implements a priority-aware scheduler working in tandem with a low-overhead preemption mechanism, enabling scheduling decisions at each layer to optimize LS task responsiveness while maintaining efficient execution flow.


\noindent \textbf{Unified sequence and batch management.} In contrast to current systems that keep concatenated batch tensors throughout response generation, a system such as QLLM requires manipulating the batch within inner layers due to context switching. Therefore, we designed a \textit{unified lifecycle management abstraction}, which encapsulates all sequence-associated states and metadata into a sequence and batch framework. This framework simulates a single batch tensor to ensure compatibility with existing models, while also allowing asynchronous updates to each sequence's individual tensors using a facade pattern abstraction.

\noindent \textbf{User-defined scheduling policies.} Tailored scheduling policies specific to the workload can optimize performance in various systems. QLLM facilitates user-defined scheduling policies (in Python) at the expert level by employing a checkpointing mechanism alongside a closed-loop controller system.

\noindent \textbf{Top-k expert selection.} Top-k expert selection for each token is an additional challenge for preemptive schedulers. To maximize QLLM’s flexibility for user-defined policies, we need a solution that enables partial processing with preemption while avoiding queue stalls due to delayed experts. Our per-expert queuing and single source of truth for state management make this efficient by pushing sequence references into multiple queues and tracking outputs as state. QLLM distinguishes fully from partially processed tokens, ensuring only complete tokens contribute to the hidden state of the next layer.

\noindent \textbf{Modular architecture.} QLLM addresses architectural bottlenecks in systems like vLLM \cite{vllm2024issue8779} by adopting a modular and extensible design. Through \textit{modularization}, \textit{encapsulation}, and \textit{dependency injection}, it decouples the system logic, minimizing interdependencies, and enabling seamless policy updates. 

\subsection{System Architecture}
\label{sec:system-architecture}

\begin{figure*}[t]
  \centering
  \includegraphics[width=\textwidth]{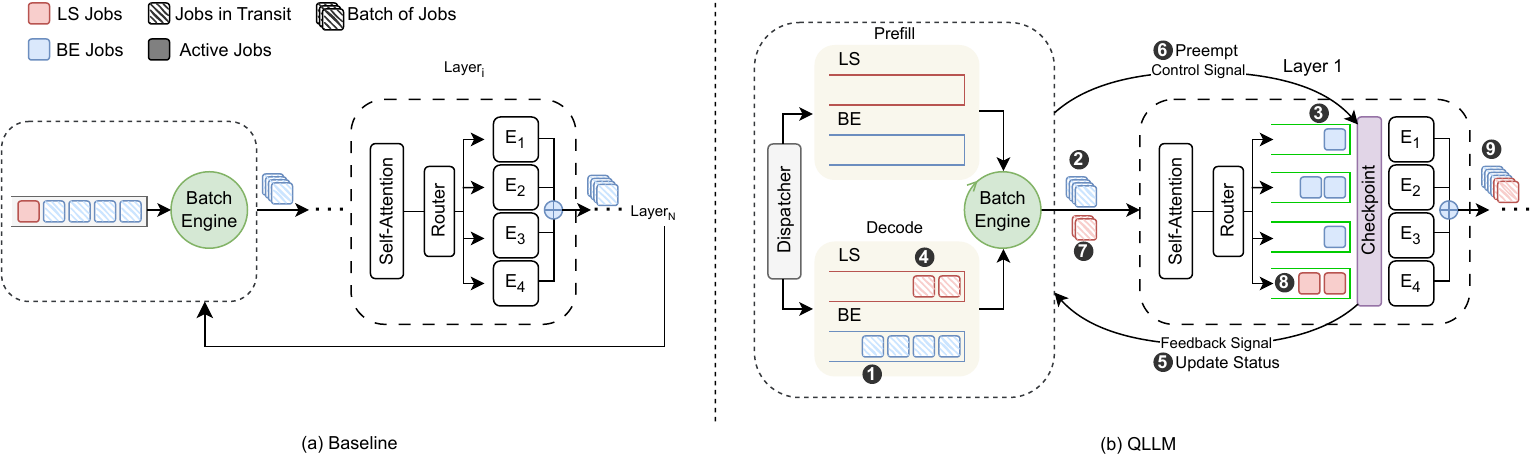}
  \caption{Comparison of a baseline and QLLM. The baseline employs iteration-level scheduling and continuous batching, with control returning to the scheduler only upon execution of all N layers. The figure on the right demonstrates a streamlined execution of QLLM's fine-grain scheduling within layer 1. LS jobs arrive after BE jobs and are batched in step 7.}
  \label{fig:system_overview}
\end{figure*}

Figure~\ref{fig:system_overview} illustrates the architecture of QLLM. At a high level, a \textit{scheduler} component receives a sequence of prompts over time—each referred to as a \textit{job}—and schedules them on the \textit{inference engine}, which processes jobs through the layers of the model. The scheduler is architected around two primary components: a \textit{dispatcher} and a \textit{batch engine}. The dispatcher enqueues incoming jobs into prefill queues based on priority and directs model output tokens into their corresponding decode queues. Simultaneously, the batch engine groups tokens into batches for each iteration, following Algorithm~\ref{alg:batch_selection}, before dispatching them to the inference engine. Unlike existing work, the QLLM scheduler enables \textit{preemption \& scheduling of jobs at fine granularity}. For example, QLLM allows preempting a batch of best-effort jobs at any processing layer to replace one of the jobs with a newly arrived latency-sensitive job, then resuming their execution. 

The QLLM scheduler utilizes four separate queues to monitor unfinished LS and BS jobs. Two of these queues manage jobs that need to be processed through the prefill stage, while the other two handle jobs that have progressed to the decode phase. Queues are implemented with an FCFS policy. The batch engine in the scheduler extracts jobs from these queues to implement a pre-configured (or a user-defined) batch policy. We show the pre-configured policy in Algorithm~\ref{alg:batch_selection} where we prioritize the execution of LS jobs by preempting BE jobs. Batches are then submitted to the inference engine.

QLLM allows for user-defined scheduling and context-switch policies at the granularity of experts, which can be implemented in fewer than 50 lines of Python code. The inference engine operates as a closed-loop feedback controller, updating the scheduler on execution status after each attention and router stage. In response, the scheduler dynamically signals the engine to adapt the execution flow based on user-defined policies. While in this paper we define our policies to minimize TTFT for LS tasks, other systems could leverage QLLM to define policies for dynamic sequence offloading or selective expert execution based on user-defined constraints, showcasing its extensibility.



\begin{algorithm}[t]\small
\caption{Batch Selection Logic}
\label{alg:batch_selection}
\begin{algorithmic}[1]
\Function{GetNextBatch}{}
    \If{LS\_DecodeQueue.size() $\geq$ BatchSize} 
        \State \Return GetBatch(LS\_DecodeQueue)

    \ElsIf{not LS\_PrefillQueue.isEmpty()} 
        \State batch $\gets$ GetBatch(LS\_PrefillQueue)
        \If{batch.size() $<$ BatchSize} 
            \State Fill from BE\_PrefillQueue
        \EndIf
        \State \Return batch

    \ElsIf{not LS\_DecodeQueue.isEmpty()} 
        \State batch $\gets$ GetBatch(LS\_DecodeQueue)
        \If{batch.size() $<$ BatchSize} 
            \State Fill from BE\_DecodeQueue
        \EndIf
        \State \Return batch

    \ElsIf{not BE\_DecodeQueue.isEmpty()} 
        \State \Return GetBatch(BE\_DecodeQueue)

    \ElsIf{not BE\_PrefillQueue.isEmpty()} 
        \State \Return GetBatch(BE\_PrefillQueue)
    \EndIf

    \State \Return None
\EndFunction
\end{algorithmic}
\end{algorithm}

\noindent \textbf{Example. } In Figure~\ref{fig:system_overview}, four BE jobs must be processed at the decode stage. Both the baseline  (Fig.~\ref{fig:system_overview}(a)) and QLLM (Fig.~\ref{fig:system_overview}(b))  batch and dispatch these jobs to the inference engine. Slightly after, an LS job arrives. With continuous batching, the baseline approach waits until all the BE jobs are processed through all the layers. Only then, it executes the prefill phase of the LS job, and  adds it to the batch. In QLLM, as soon as an LS job arrives, the scheduler stops the execution of the BE batch at the engine. It then executes the prefill stage for the LS job, and then the subsequent decode stage. Then, QLLM dynamically merges the LS jobs with the BE jobs within the layer execution, reducing the latency of the LS job. 

\noindent \textbf{Batch selection logic.} Now we explain how Algorithm~\ref{alg:batch_selection} creates batches. The goal is to fill a batch with LS jobs without exceeding the maximum batch size. The algorithm prioritizes filling the batch first with LS jobs at the decode stage (lines 2-3). If a maximum-sized batch cannot be created, it prioritizes LS jobs in the prefill stage to produce more LS jobs in the decode stage (lines 4-5). The algorithm also attempts to add some BE jobs to fill the batch if possible (lines 6-7). If there are no LS jobs at the prefill stage, the algorithm simply executes the LS jobs at the decode stage, filling the batch with BE jobs at the decode stage (lines 10-14). If there are no LS jobs, it simply executes BE jobs, prioritizing decode over prefill (lines 16-19).

\subsection{System Modularity and Extensiblity} 

\noindent \textbf{Unified sequence and batch abstractions for inference.}
At its core, QLLM replaces the fragmented sequence and batch handling found in existing inference systems with a unified execution model. Rather than managing sequences and their state in an ad hoc manner, QLLM defines a Sequence abstraction, encapsulating all relevant metadata, including KV cache tensors, routing information, and execution state. A corresponding Batch abstraction allows collective processing while maintaining individual sequence integrity. This structured approach streamlines execution, improves observability, and ensures robust preemption without introducing unnecessary synchronization overhead.  

\noindent \textbf{Breaking rigid batch processing with queues per expert. }
A major advancement is the per-expert FIFO queuing, which fundamentally transforms the token flow through MoE layers. In contrast to traditional inference engines that create rigid batch tensors retained across iterations until execution completion, QLLM employs distinct and independent data structures-including tensors for each Sequence object-and offers a unified interface through abstraction, ensuring compatibility without necessitating destructive modifications.

This is achieved through the \textit{Facade Pattern}, where the \textit{Batch} abstraction acts as an interface between the model and underlying sequence-level data structures. Instead of handling anonymous concatenated tensors, QLLM provides a structured representation where each \textit{Sequence} object retains its own state while being processed within a unified \textit{Batch}. This allows the model to interact with monolithic tensors while the underlying system dynamically tracks and updates individual sequences upon tensor modification. By intercepting tensor access, QLLM enables fine-grained control over execution flow, facilitating expert-level preemption and efficient context switching without modifying core model operations.

\noindent \textbf{Eliminating costly split-merge procedures.}
Unlike existing commercial systems where retrieving a token from a batch involves costly splitting and concatenating of pre-constructed tensors, QLLM enables direct, real-time state updates of individual sequences, facilitating low-overhead preemption by overlapping token state updates with execution. Additionally, QLLM introduces efficient KV cache management through a novel module which we call the Unified Dynamic Cache, which decouples sequence-level and batch-level cache operations and avoids expensive split-merge procedures on large KV tensors.

\noindent \textbf{Implications for deferred inference execution. } In addition to preemption, per-expert queuing opens up possibilities for researchers to investigate optimizations in deferred execution, adaptive load balancing, dynamic workload distribution, fault tolerance, multi-tenant environments, and beyond. While QLLM is centered on low-latency inference for mixed-priority workloads, its foundational architectures are applicable to future systems demanding enhanced flexibility in MoE inference. Nevertheless, the concepts of per-expert queuing and low-overhead state management are not confined to MoE models and may be applied to dense models.




\section{Evaluation}

This section offers an initial evaluation of QLLM's performance utilizing practical hardware configurations, with an emphasis on its impact on the latency of LS requests and the overall turnaround time. We address the following questions:
\begin{itemize}[leftmargin=*,noitemsep]
    \item \textbf{Q1:} How does QLLM comply with the Latency SLO?
    \item \textbf{Q2:} How does QLLM affect throughput?
    \item \textbf{Q3:} How does QLLM affect the turnaround time for BE and LS requests?
\end{itemize}

\noindent \textbf{Experimental Setup.}
The evaluation was conducted on a bare-metal system equipped with an Nvidia A100 GPU (80~GB memory), dual-socket Intel Xeon Gold 6336Y CPUs, 256~GB DRAM, and PCIe 4.0 interconnect.

\noindent \textbf{Models and Dataset.}
We use Mixtral 8×7B, a representative of MoE models. The model is executed with 4-bit quantization and FP16 precision. In this configuration, the model required approximately 22.93 GB of GPU memory. Experiments are conducted using the ShareGPT dataset~\cite{sharegpt}.

\noindent \textbf{Baseline System.}
To evaluate the performance of QLLM, we compare it against the HF TGI inference engine, which represents a widely used production engine in serving LLM models. Importantly, QLLM is built on top of the HF engine, ensuring that any performance gains can be attributed to the scheduling mechanism rather than unrelated system differences. We set the maximum batch size to 32.

\noindent \textbf{Workload.}
The workload generator retrieves prompts from the ShareGPT dataset and marks 20\% of these prompts as LS prompts based on a random selection process. Subsequently, requests are dispatched to QLLM in accordance with Poisson arrival rates. It is important to know that QLLM performance may vary under different workload patterns, including the portion of LS requests and their distribution. However, in this preliminary evaluation of our prototype system, we explore the workload described earlier.

\noindent \textbf{Evaluation Metrics.} 
For users, the latency of LS requests significantly influences the responsiveness of applications like code completion and medical LLM applications. Therefore, we evaluate and present both TTFT and the turnaround time for LS jobs. Turnaround time refers to the complete delay from the moment a request is received by the system until the response is fully generated and delivered to the user. Furthermore, we measure the \textit{job completion rate} as a throughput metric. Additionally, we investigate the impact of QLLM on the BE turnaround time.

\begin{figure}[t]
  \centering
  \includegraphics[width=\columnwidth]{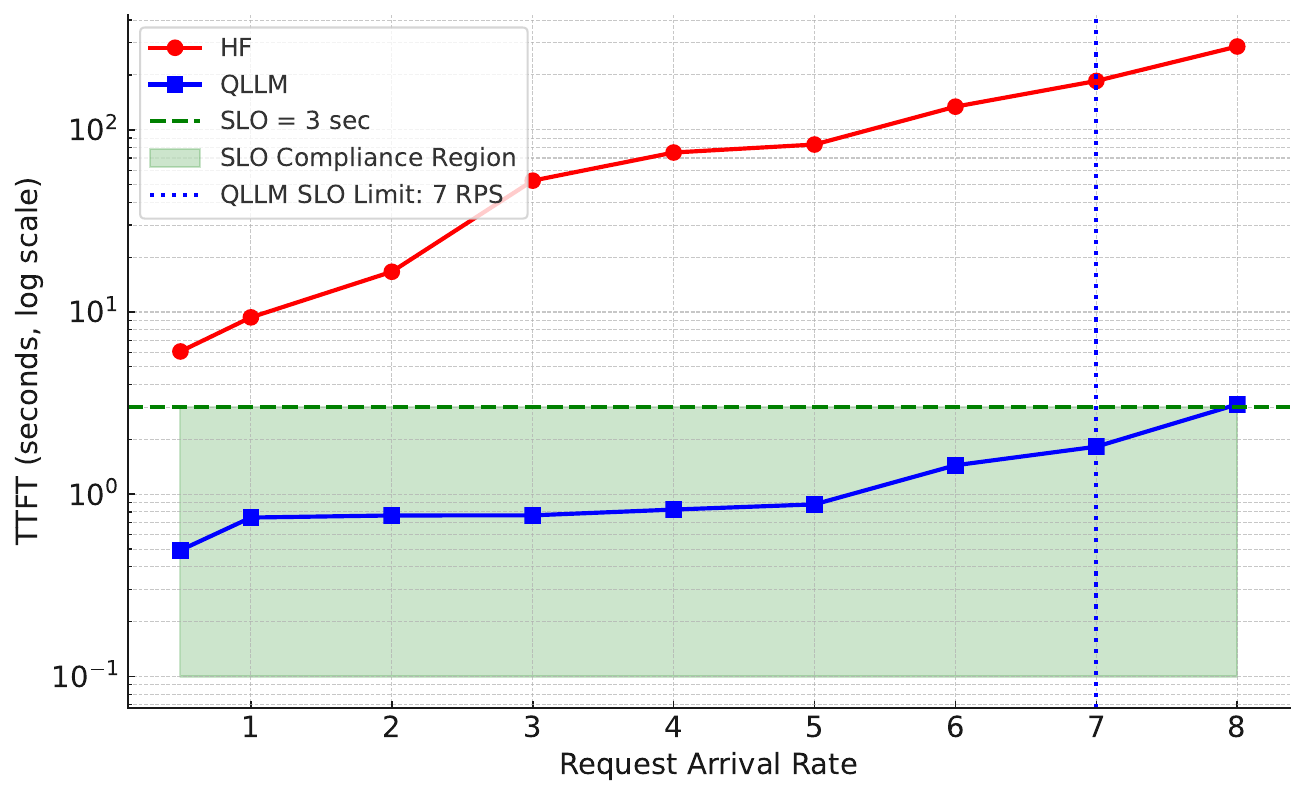}
  \caption{QLLM reduces TTFT for LS jobs by up to $101.6\times$ while ensuring compliance with the SLO. In contrast, Hugging Face fails to meet the SLO even under low load due to priority-oblivious scheduling.}
  \label{fig:avg-ls-ttft}
\end{figure}

\vspace{.05in}
\noindent \textbf{Q1: QLLM significantly decreases TTFT latency.}
Figure~\ref{fig:avg-ls-ttft} compares the TTFT of the HF TGI  baseline (red line) with QLLM (blue line). We set the SLO to 3 seconds (green dashed line), which is $10\times$ the processing iteration time of a single decode \cite{wu2023fastserve, Prekas2017ZygOS}. The results show that the priority-oblivious scheduler of HF TGI cannot comply with the SLO even at low request rates. In fact, HF TGI employs a run-to-completion strategy in which lengthy BE jobs postpone the execution of LS jobs. In contrast, QLLM handles up to $7$ jobs\slash s  while still adhering to the SLO latency for LS requests. Notably, QLLM reduces TTFT by up to $101.6\times$ and an average of $65.2\times$ while complying with the SLO.

\begin{figure}[t]
  \centering
  \includegraphics[width=\columnwidth]{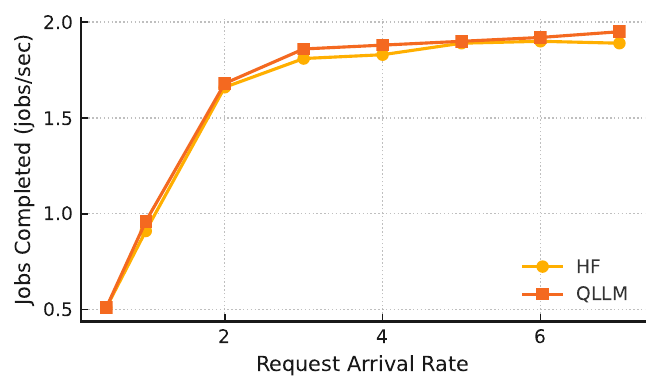}
  \caption{QLLM maintains a comparable or slightly higher job completion rate compared to HF.}
  \label{fig:job_completion_rate}
\end{figure}

\vspace{.05in}
\noindent \textbf{Q2: QLLM preserves overall throughput.}
Figure~\ref{fig:job_completion_rate} depicts the variation in throughput as a function of the request arrival rates, quantified by the job completion metric. QLLM demonstrates throughput that is comparable to, or slightly exceeds, the baseline while adhering to the latency SLO for LS requests. Its ability to execute fine-grained preemption enables QLLM to promptly address incoming LS requests while simultaneously managing best-effort decoding jobs, thereby increasing GPU efficiency. The throughput measured in tokens per second exhibits a similar trend.

\begin{figure}[t]
\centering
\begin{minipage}[c]{0.49\columnwidth}
\centering
\includegraphics[width=\columnwidth]{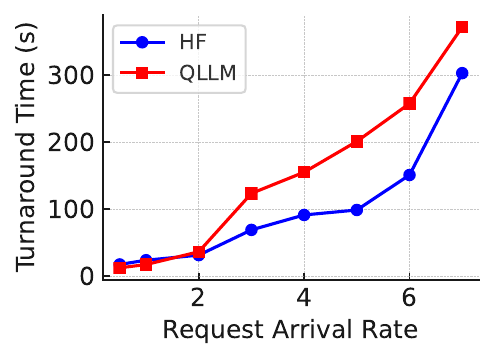}
\subcaption{BE Requests}
\label{fig:be-turnaround-time}
\end{minipage}
\hfill
\begin{minipage}[c]{0.49\columnwidth}
\centering
\includegraphics[width=\columnwidth]{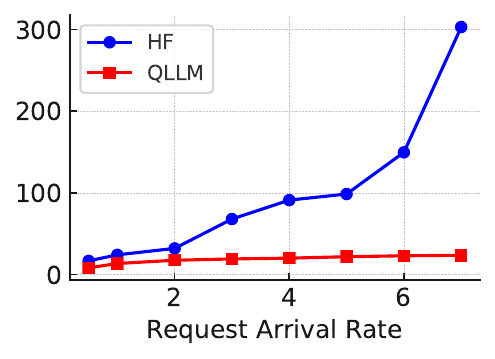}
\subcaption{LS Requests}
\label{fig:ls-turnaround-time}
\end{minipage}
\caption{Comparison of turnaround times for Best Effort and Latency-Sensitive requests.}
\label{fig:turnaround-time}
\end{figure}

\vspace{.05in}
\noindent \textbf{Q3: QLLM reduces turnaround time for LS jobs.}
Figure~\ref{fig:turnaround-time}  compares the turnaround time of HF TGI (blue line) and QLLM (red line) for BE and LS jobs. 
For LS jobs, QLLM reduces up to $12.8\times$ the turnaround time relative to the baseline thanks to its preemption and low-latency response to LS requests. 
QLLM experiences a $1.38\times$ increase in response time for BE requests, reaching a peak of $2.04\times$.
The findings suggest that the QLLM advantages for LS jobs considerably surpass the detriments to BE requests. 

\section{Related Work}
\label{sec:discussion}

Existing approaches to LLM inference scheduling primarily optimize at the iteration level, lacking finer-grained control. ORCA \cite{jeon2023orca} and FastServe \cite{wu2023fastserve} optimize scheduling at the iteration level, primarily benefiting dense models but lacking finer-grained control. Although FastServe introduces token-level preemption, it does not incorporate priority-aware execution. Additionally, Reef \cite{reef} focuses on microsecond-scale preemption for traditional DNN model serving, where each request involves a single inference pass through the model. In contrast, LLM inference operates autoregressively, requiring multiple iterations.

Llumnix \cite{sun2024llumnix} and FastSwitch \cite{shen2024fastswitch} enhance scheduling by managing KV cache migration and preemptive context switching, respectively, but still operate within iteration-level scheduling. 
Unlike these approaches, our work achieves more fine-grained control, enabling efficient priority management and dynamic execution without excessive recomputation overhead.

In contrast, QLLM enables expert-level preemption and priority-aware scheduling, addressing head-of-line blocking and priority inversion. This fine-grained control ensures low latency for LS tasks while maintaining high throughput, surpassing prior iteration-level approaches.

\section{Discussion}

\noindent \textbf{Limitations.} While the existing QLLM prototype demonstrates promising results, we are actively exploring methods to reduce memory overhead and mitigate potential starvation in certain workloads. Compared to iteration-level preemption, our approach requires caching additional states (\eg, \textit{routing\_weights} and \textit{hidden\_states}), though the primary memory constraint remains the KV cache. Efficient memory management to further enhancing the effectiveness of preemptive scheduling could be a subject for future work.

\noindent\textbf{Overlapping preemption with execution.} The new MoE layer introduces execution flexibility, allowing the system to process entire batches, selectively execute specific tokens or experts, or dynamically preempt ongoing tasks. This adaptability enables opportunities for overlapping memory operations with concurrent task execution, which can improve overall performance. Exploring these optimizations can further enhance system efficiency.  

\noindent \textbf{Applicability to dense models.} Our approach extends beyond MoE models and is applicable to all LLM architectures. In dense models, preemptive scheduling at the layer level is more straightforward due to their deterministic execution, where all tokens follow the same computational path. However, MoE models introduce dynamic token-to-expert routing, necessitating more sophisticated state management and preemption mechanisms. By addressing these complexities, QLLM provides a generalized scheduling framework that supports both MoE and dense models.

QLLM’s architectural flexibility introduces new opportunities for optimizing LLM inference, enabling more efficient scheduling and execution strategies. This design opens pathways for further exploration in adaptive workload management, memory-efficient preemption, and broader applications beyond MoE models. By redefining how inference tasks are scheduled and executed, QLLM lays the groundwork for future advancements in efficient and scalable LLM serving.

\vspace{-2pt}

\section{Conclusion}

This paper introduces QLLM, the first inference system that facilitates fine-grained preemption and priority-aware scheduling for MoE models, optimizing latency-sensitive jobs while preserving high throughput. By implementing per-expert queues and a priority-aware scheduler, QLLM addresses HOL blocking and priority inversion, achieving a reduction in LS jobs TTFT latency by up to $101.6\times$ and ensuring SLO compliance up to 7 requests/sec, whereas the baseline never adheres to SLO in tested scenarios. The proposed approach is modular and extensible, allowing seamless integration with existing MoE frameworks. Future research will focus on further optimizing memory consumption, potential starvation, and open-source QLLM to drive continued innovation in efficient LLM inference.

\vspace{-4pt}
\section*{Acknowledgments}

We would like to thank the anonymous reviewers for their insightful comments and suggestions on this paper. This work has been partially supported by Vinnova (the Sweden's Innovation Agency), the Swedish Research Council (agreement No. 2021-04212), KTH Digital Futures, and Knut and Alice Wallenberg Foundation (Wallenberg Scholar Grant for Prof. Dejan Kosti\'c). 


\bibliographystyle{ACM-Reference-Format}
\bibliography{sample-base}





\end{document}